%% file: baidukddcup2022.tex
\documentclass[sigconf]{acmart}

\AtBeginDocument{%
  \providecommand\BibTeX{{%
    \normalfont B\kern-0.5em{\scshape i\kern-0.25em b}\kern-0.8em\TeX}}}

\newcommand{\rev}[1]{\textcolor{black}{#1}}

\newcommand{\polish}[1]{\textcolor{black}{#1}}

\newcommand{\revtwo}[1]{\textcolor{black}{#1}}

\acmConference[Baidu KDD Cup 2022]{The 28th ACM SIGKDD
Conference on Knowledge Discovery and Data Mining}{Mar. 16 -- Jul. 17, 2022}{}
\begin{document}

\title{
SDWPF: A Dataset for Spatial Dynamic Wind Power Forecasting Challenge at KDD Cup 2022
}


\author{
Jingbo Zhou$^{1*}$, Xinjiang Lu$^{1*}$, Yixiong Xiao$^{1*}$, Jiantao Su$^3$, Junfu Lyu$^4$, Yanjun Ma$^2$, Dejing Dou$^1$
}\thanks{$^*$ Equal contribution.}
\affiliation{%
  \institution{$^1$Baidu Research  $^2$Baidu Inc. $^3$ Longyuan Power Group Corp. Ltd., $^4$Tsinghua University}
  \city{}
  \country{}
}
  
\affiliation{%
\institution{\{zhoujingbo, luxinjiang, xiaoyixiong, mayanjun02, doudejing\}@baidu.com}
\institution{$^3$12091329@chnenergy.com.cn,
$^4$lvjf@mail.tsinghua.edu.cn, 
}
 \city{}
  \country{}
}

\renewcommand{\shortauthors}{Zhou, et al.}

\begin{abstract}
The variability of wind power supply can present substantial challenges to incorporating wind power into a grid system.  Thus, Wind Power Forecasting (WPF) has been widely recognized as one of the most critical issues in wind power integration and operation. There has been an explosion of studies on wind power forecasting problems in the past decades. Nevertheless, how to well handle the WPF problem is still challenging, since high prediction accuracy is always demanded to ensure grid stability and security of supply. We present a unique Spatial Dynamic Wind Power Forecasting dataset: SDWPF, which includes the spatial distribution of wind turbines, as well as the dynamic context factors. 
Whereas, most of the existing datasets have only a small number of wind turbines without knowing the locations and context information of wind turbines \rev{at} a fine-grained time scale. 
By contrast, SDWPF provides the wind power data of 134 wind turbines from a wind farm over half a year with their relative \rev{positions} and internal \rev{statuses}.
We use this dataset to launch the Baidu KDD Cup 2022 to examine the limit of current WPF solutions. 
\rev{The dataset is released at}
\url{https://aistudio.baidu.com/aistudio/competition/detail/152/0/datasets}. 
\end{abstract}


\maketitle

\input{section/intro}
\input{section/related}
\input{section/dataset}

\input{section/evaluation}

\input{section/exp}

\bibliographystyle{ACM-Reference-Format}
\bibliography{ref}
\end{document}

%% file: section/intro.tex
\section{Introduction}

Wind Power Forecasting (WPF) aims to accurately estimate the wind power supply of a wind farm at different time scales. Wind power is a kind of clean and safe source of renewable energy, but cannot be produced consistently, leading to high variability. Such variability can present substantial challenges to incorporating wind power into a grid system. To maintain the balance between electricity generation and consumption, the fluctuation of wind power requires power substitution from other sources that might not be available at short notice (for example, usually it takes at least 6 hours to fire up a coal plant). Thus, WPF has been widely recognized as one of the most critical issues in wind power integration and operation. There has been an explosion of studies on wind power forecasting problems appearing in the data mining and machine learning community. Nevertheless, how to well handle the WPF problem is still challenging, since high prediction accuracy is always demanded to ensure grid stability and security of supply.

We present a unique Spatial Dynamic Wind Power Forecasting dataset: SDWPF, which includes the spatial distribution of wind turbines, as well as the dynamic context factors like \polish{temperature}, weather, and turbine internal status. Whereas, many existing datasets and competitions treat WPF as a time series prediction problem without knowing the locations and context information of wind turbines.

SDWPF is obtained from the real-world data from Longyuan Power Group Corp. Ltd. (the largest wind power producer in China and Asia). There are two unique features for this competition task different from previous WPF competition settings: 1) Spatial distribution: this competition provides the relative location of all wind turbines given a wind farm for modeling the spatial correlation among wind turbines. 2) Dynamic context: the weather situations and turbine internal status detected by each wind turbine are provided to facilitate the forecasting task.

With aiming to examine the limit of WFP methods, we use 
the SDWPF dataset to launch the Baidu KDD Cup 2022 Challenge.  SDWPF contains the wind power data obtained from the Supervisory Control And Data Acquisition (SCADA) system of a wind farm which has 134 wind turbines. The dataset provide the information about the wind, temperature, turbine angle and historical wind power. The time range of the dataset is over half a year. We also provide a baseline for this dataset\footnote{\url{https://github.com/PaddlePaddle/PaddleSpatial/tree/main/apps/wpf_baseline_gru}}. The introduction about the challenge can found in the Baidu KDD Cup 2022 website\footnote{\url{https://aistudio.baidu.com/aistudio/competition/detail/152/0/introduction}} and the dataset can be down after registration\footnote{\url{https://aistudio.baidu.com/aistudio/competition/detail/152/0/datasets}}. 

%% file: section/related.tex
\section{Related Work}

Wind power forecasting (WPF) has been extensively investigated over the past decades \cite{wang2011review,foley2012current,sideratos2007advanced,deng2020wind}. According to the spatial scale of the wind power, the problem can be categorised as a single wind turbine, a wind farm and a group of wind farms \cite{jiang2019review}. The dataset of this challenge belongs to the wind farm scale. A few of delicate models have been specially designed for WPF problem with variant of spatial and temporal scales based on statistic models \cite{sideratos2007advanced,milligan2003statistical}, machine learning methods \cite{zeng2011support,hu2015short} and deep learning methods \cite{wang2017deep, hong2019hybrid}. Many advanced time series prediction methods like \cite{zhou2015smiler,liang2018geoman,hu2020multistage,li2020autost,zhou2021informer,wu2021autoformer} also have great potential to tackle this problem. 

Though there are a few of public WPF datasets, they usually have only a \polish{limited} number of wind \polish{turbines} \polish{and do not provide} the spatial information of \polish{each} turbine. For example, the Penmanshiel dataset has only 14 turbines \cite{penmanshiel2022}, and the Kaggle dataset has only 1 turbine \cite{kaggle2022}.  We leave a comprehensive discussion about the WPF methods and WPF datasets as a future work.

%% file: section/dataset.tex
\section{SDWPF Dataset}
In this section, we provide an brief introduction  of the SDWPF dataset, including data source, overall statistics, schema and spatial distribution. The SDWPF dataset is collected from \polish{the} Supervisory Control and Data Acquisition (SCADA) \polish{system of} a wind farm. The SCADA data are sampled \textbf{every 10 minutes} from each wind turbine in the wind farm which consists of 134 wind turbines. The statistics of the important information of the SDWPF dataset is shown in the Table \ref{table-stat}.

\begin{table}[htb]
\centering
\begin{tabular}{ccccc}
	\toprule
		Days & Interval &  \# of columns & \# of turbines &\# of records\\
		\midrule
		\polish{245}	& 10 minutes& 13  & 134 & \polish{4,727,520}	\\
		\bottomrule
\end{tabular}
\caption{Statistics of the SDWPF data.} \label{table-stat}
\end{table}

The dataset includes critical external features, such as wind speed, wind direction and external temperature, that influence the wind power generation; as well as essential internal features, such as the inside temperature, nacelle direction and Pitch angle of blades, which can indicate the \polish{operating} status of each wind turbine.

\begin{figure}[t]
\centering
\includegraphics[width=0.7\columnwidth]{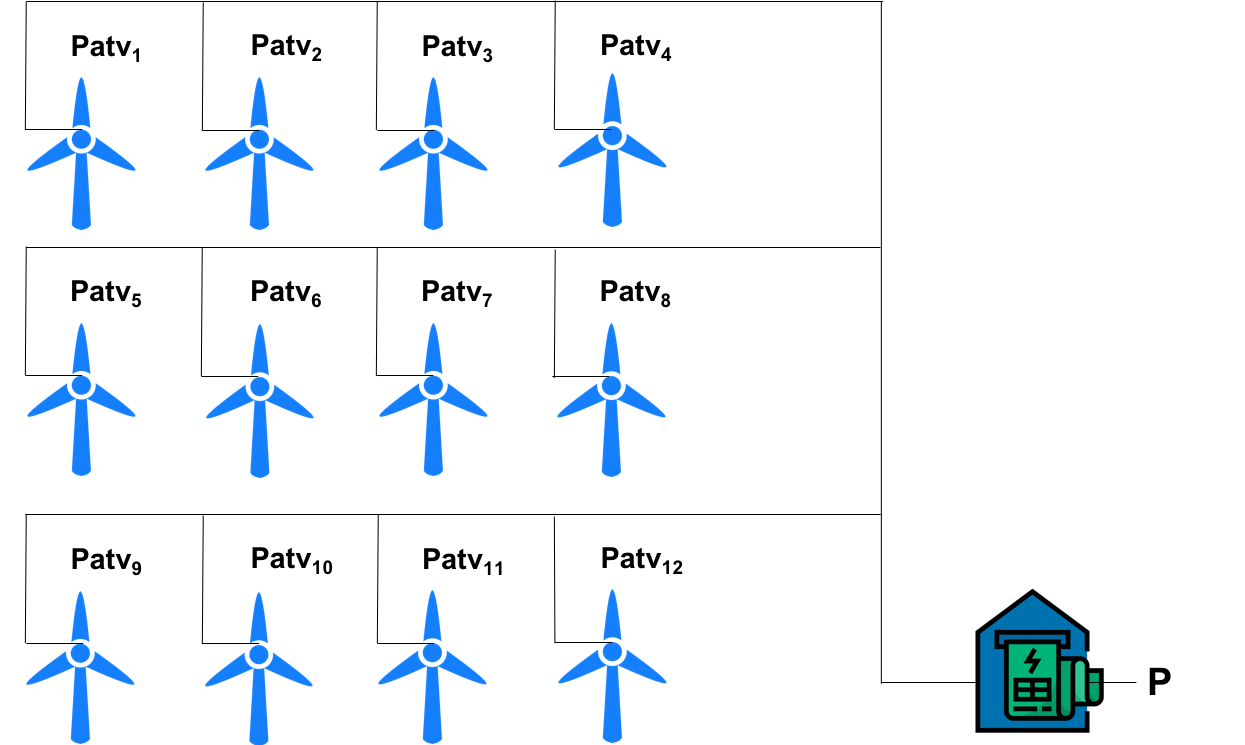}
\caption{An illustration of a wind farm.}
\label{fig-illus}
\end{figure}

Each wind turbine can generate the wind power $Patv^i$ separately, and the outcome power of the wind farm is the sum of all the wind turbines. 
In other words, at time \polish{\textit{t}}, the output power of the wind farm is $P=\sum_i Patv^i$. An illustration of a wind farm is shown in Figure \ref{fig-illus}.  
We also provide a detailed introduction about the main attributes of the data in Table \ref{table-schema}. 
Please refer to Wikipedia for more details about  components of wind turbines\footnote{
We suggest to refer to 
\url{https://en.wikipedia.org/wiki/Wind\_turbine\#Horizontal\_axis} and   
\url{https://en.wikipedia.org/wiki/Wind\_turbine\#Components}
}.

\begin{table*}[htb]
	\centering
	\begin{tabular}{ccl}
		\toprule
		Column	& Column Name	& Specification \\
		\midrule
		1	& TurbID  & Wind turbine ID		\\
		2	& Day  & {Day of the record}	\\
		3&	Tmstamp	&Created time of the record \\
		4&	Wspd (m/s)&	The wind speed recorded by the anemometer \\
		5&	Wdir(°)	&The angle between the wind direction and the position of turbine nacelle\\
		6&	Etmp (°C)&	Temperature of the surounding environment \\
		7&	Itmp (°C)&	Temperature inside the turbine nacelle \\
		8&	Ndir (°)&	Nacelle direction, i.e., the yaw angle of the nacelle\\
		9&	Pab1 (°)&	Pitch angle of blade 1  \\
		10&	Pab2 (°)&	Pitch angle of blade 2  \\
		11&	Pab3 (°)&	Pitch angle of blade 3  \\
		12&	Prtv (kW)&	Reactive power \\
		13&	Patv (kW)&	Active power (target variable)\\
		\bottomrule
	\end{tabular}
		\caption{Column names and their specifications of the SDWPF data.}
	\label{table-schema}
\end{table*}

\begin{figure}[t]
\centering
\includegraphics[width=0.65\columnwidth]{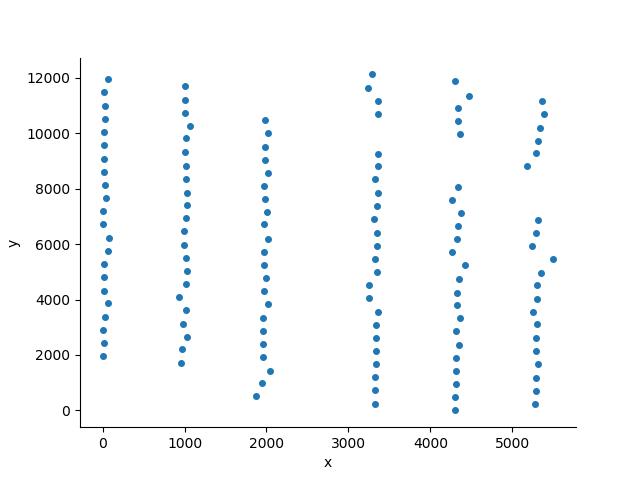}
\caption{Spatial distribution of all wind turbines (x and y are with the meter unit).}
\label{fig-turbine_position}
\end{figure}

The relative position of all wind turbines in the wind farm is also released to characterize the spatial correlation between wind turbines. An illustration of the spatial distribution of the totally 134 wind turbines are shown in Figure \ref{fig-turbine_position}. The units of x and y are meter.

%% file: section/evaluation.tex
\section{Evaluation}

The Baidu KDD Cup 2022 requires to address the Spatial Dynamic Wind Power Forecasting ahead of 48 hours. 
For example, given at 6:00 A.M. today, it is required to effectively forecast the wind power generation beginning from 6:00 A.M. on this day to 5:50 AM \rev{on the day after tomorrow}, given a series of historical \rev{records} of the wind farm \rev{and the related} wind turbines. 
It is required to output the predicted values every 10 minutes. 
To be specific, at one \rev{time point}, it is required to predict a future length-288 wind power supply time-series. 
The average of RMSE (Root Mean Square Error) and MAE (Mean Absolute Error) is used as the main evaluation score.  

Note in our settings, we aim to forecast the power generated by a wind farm with the SCADA data and spatial data on top of the spatiotemporal modeling paradigm without knowing the future meteorological data (wind speed, temperature, etc.).
During the Baidu KDD Cup 2022 challenge, except \revtwo{the released data of 245 days}, we still privately hold data of several months to evaluate the submitted models by participants. Before giving a formal definition of the metric, we first present some caveats about the data.
 
\subsection{Caveats about the data}

Here we introduce a few of caveats when to use this data to train and evaluate the models.
 
 \textbf{Zero values}. There are some active power and reactive power which are smaller than zeros. We simply treat all the values which are smaller than 0 as 0, i.e. if $Patv<0$, then $Patv=0$.

 \textbf{Missing values}. Note that due to some reasons, some values at some time are not collected from the SCADA system. These missing values will not be used for evaluating the model. In other word, if $p_{t_0+j}$ is a missing value, we set $|Patv_{t_0+j}-\overline{Patv}_{t_0+j}|=0$ regardless of the actual predicted value of $\overline{Patv}_{t_0+j}$.
 
 \textbf{Unknown values}. In some time, the wind turbines are stopped to generate power by external reasons such as wind turbine renovation and/or actively scheduling the powering to avoid overloading the grid. In these cases, the actual generated power of the wind turbine is unknown. These unknown values will also not be used for evaluating the model. Similarly with the missing values, if $Patv_{t_0+j}$ is a unknown value, we always set $|Patv_{t_0+j}-\overline{Patv}_{t_0+j}|=0$. Here we introduce two conditions to determine whether the target variable is unknown:
 \begin{itemize}
     \item If at time $t$, $Patv\leq 0$ and $Wspd>2.5$, then the actual active power $Patv$ of this wind turbine at time $t$ is unknown;
     \item If at time $t$, $Pab1>89^{\circ}$ or $Pab2>89^{\circ}$ or $Pab3>89^{\circ}$, then the actual active power $Patv$ of this wind turbine at time $t$ is unknown.
 \end{itemize}
 
\revtwo{
 \textbf{Abnormal values}
 There are some abnormal values from the SCADA system. If a data record has any abnormal value of any column, these values also will not be used for evaluating the model. If $Patv_{t_0+j}$ is a abnormal value, we always set $|Patv_{t_0+j}-\overline{Patv}_{t_0+j}|=0$.
 Here we define two rules to identify the abnormal values:
 \begin{itemize}
     \item The reasonable range for Ndir is [-720°, 720°], as the turbine system allows the nacelle to turn at most two rounds in one direction and would force the nacelle to return to the original position otherwise. Therefore, records beyond the range can be seen as outliers caused by the recording system. Thus, if at time $t$ there are Nidir > 720° or Nidir < -720°, then the actual active power $Patv$ of this wind turbine at time $t$ is abnormal.
     \item The reasonable range for Wdir is [-180°, 180°]. Records beyond this range can be seen as outliers caused by the recording system. If at time $t$ there are Widr > 180° or Widr < -180°, then the actual active power $Patv$ of this wind turbine at time $t$ is abnormal.
 \end{itemize}
}

\subsection{Evaluation metrics}

\revtwo{Formally, at a time step  $t_0$, it is required to predict a time series of wind power of the wind farm 
\revtwo{
$P=\{p_{t_0+1}, p_{t_0+2}, \cdots , p_{t_0+288}\}$
}. However, due to the missing and unknown values for each wind turbine, in this challenge, we evaluate the prediction results for each wind turbine, and then sum the prediction scores as the final score of the model. The evaluation score $s^i_{t_0}$for wind turbine $i$ at the time step $t_0$ is defined as:}

\revtwo{
\begin{equation}
    s^i_{t_0} = \frac{1}{2}(\sqrt{\frac{\sum_{j=1}^{288}(Patv^i_{t_0+j}-\overline{Patv}^i_{t_0+j})^2}{288}} + \frac{\sum_{j=1}^{288}|Patv^i_{t_0+j}-\overline{Patv}^i_{t_0+j}|}{288})
\end{equation}
}

\noindent 
\revtwo{
where $Patv^i_{t_0+j}$  is the actual power of wind turbine $i$ and $\overline{Patv}^i_{t_0+j}$  is the predicted power of the wind turbine $i$ at time step $t_0+j$. Note that each time step of $j$ is 10 minutes. The overall score of the prediction model $S_{t_0}$ at time $t_0$ is the sum of the prediction score on all wind turbine, i.e.:
\begin{equation}
S_{t_0} = \sum_{i=1}^{134} s^i_{t_0}
\end{equation}
}

\revtwo{
A length-$L_x$-length-288 prediction window is adopted to roll the whole test set with stride $\Delta t$ time steps (Each time step of $\Delta t$ is 10 minutes), and the averaged evaluation score is reported. 
Note that, $L_x$ denotes the length of input time-series.
}

\revtwo{
We use $\mathbf{K}$ data instances to evaluate the performance of the prediction model. 
For each data instances $k$, we randomly sample a stride time step $\Delta t_k$ from the range 
[1, 10]. 
In other word, the stride time step are randomly ranged from 
10 minutes to 100 minutes. 
Formally, the evaluation score of the model is:
}

\revtwo{
\begin{equation}\label{eqn:score}
 score = \frac{1}{K} \sum_{k=0}^{\mathbf{K}} S_{t_0 + \sum_{r=0}^{k} \Delta t_r}
\end{equation}
}

\noindent
\revtwo{
The code to calculate the score is available in our sample code.
}
\revtwo{Here we would like to highlight the following important points:
\begin{itemize}
\item In our evaluation, we set the maximum length of the input time series $L_x$ as 14 days.
\item As shown in our sample code, since we sum the error from all wind turbines, in order to avoid the large value, we use the Mega Watt (instead of Kilo Watt) as the unit to represent the final score.
\item For the evaluation in our submission system (starting from May 10), we use 195 random sampled stride time steps (i.e. $\{\Delta t_0, \Delta t_1, ... ,  \Delta t_{194}\}$ in Eqn. \ref{eqn:score})  over several months to evaluate the submitted models. This test time range and sampled stride time steps will be updated in future phases (June 20 and July 15).
\end{itemize}
}


%% file: section/exp.tex
\section{Baseline code}
We have released a simple baseline code with Gated Recurrent Unit \cite{cho2014properties} in PaddleSpatial\footnote{\url{https://github.com/PaddlePaddle/PaddleSpatial/tree/main/apps/wpf_baseline_gru}}.
In our experiment, we sum the active power of all the wind turbines to form a wind power time series. 
And then we use the data of first 214 days as training data and the remain 31 days as validation data. 
According to our statistics, 
\revtwo{
the score of our baseline over the tested time series of 195  predictions (i.e. $K=195$) is: RMSE: 47.081286, MAE: 37.558233, and the overall score is 42.319760.
%
The evaluation time for 195 predictions is 1129.722821 seconds on a Linux machine with Nvidia P40 GPU. 
Note that, the batch size impacts the performance evaluation. 
To alleviate this, the evaluation adopts the strategy without batching and tests the performance on one instance at a time. 
}